\documentclass[10pt, a4paper]{article}
\usepackage{lrec}
\usepackage{graphicx}
\usepackage{tabularx}
\usepackage{listings}
\usepackage{soul}
\usepackage{epstopdf}
\usepackage[utf8]{inputenc}
\usepackage[hidelinks]{hyperref}
\usepackage{xstring}
\usepackage{xcolor}
\usepackage{lipsum}

\newcommand{\secref}[1]{\StrSubstitute{\getrefnumber{#1}}{.}{}}

\title{Making Metadata Fit for Next Generation Language Technology Platforms: \\ The Metadata Schema of the European Language Grid}

\name{Penny Labropoulou\textsuperscript{1}, Katerina Gkirtzou\textsuperscript{1}, Maria Gavriilidou\textsuperscript{1}, Miltos Deligiannis\textsuperscript{1},  Dimitrios \\ 
\large\bf{Galanis\textsuperscript{1}, Stelios Piperidis\textsuperscript{1}, Georg Rehm\textsuperscript{2}, Maria Berger\textsuperscript{2}, Valérie Mapelli\textsuperscript{3}, Mickaël Rigault\textsuperscript{3},}\\ 
\large\bf{Victoria Arranz\textsuperscript{3}, Khalid Choukri\textsuperscript{3}, Gerhard Backfried\textsuperscript{4}, José Manuel Gómez Pérez\textsuperscript{5},}\\ \large\textbf{Andres Garcia Silva\textsuperscript{5}}}

\address{
        \textsuperscript{1}ILSP/Athena RC, Greece, \textsuperscript{2}DFKI GmbH, Germany, \textsuperscript{3}Evaluations and Language Resources Distribution Agency (ELDA), \\ France, \textsuperscript{4}SAIL LABS Technology GmbH, Austria, \textsuperscript{5}Expert System Iberia SL, Spain\\
         \{penny, gkirtzou, maria, mdel, galanisd, spip\}@athenarc.gr, \{georg.rehm, maria.berger\}@dfki.de, \\ 
         \{mapelli, mickael, arranz, choukri\}@elda.org, Gerhard.Backfried@sail-labs.com, \{agarcia, jmgomez\}@expertsystem.com}

\abstract{The current scientific and technological landscape is characterised by the increasing availability of data resources and processing tools and services. In this setting, metadata have emerged as a key factor facilitating management, sharing and usage of such digital assets. In this paper we present ELG-SHARE, a rich metadata schema catering for the description of Language Resources and Technologies (processing and generation services and tools, models, corpora, term lists, etc.), as well as related entities (e.g., organizations, projects, supporting documents, etc.). The schema powers the European Language Grid platform that aims to be the primary hub and marketplace for industry-relevant Language Technology in Europe. ELG-SHARE has been based on various metadata schemas, vocabularies, and ontologies, as well as related recommendations and guidelines.  \newline 
\Keywords{metadata, language technology, language technology services, language resources}}

\begin{document}

\maketitleabstract

\section{Introduction} 
\label{sec:introduction}

The rise of data-driven approaches that use Machine Learning (ML), and especially the breakthroughs in the Deep Learning field, has put data into a central place in all scientific and technological areas, Natural Language Processing (NLP) being no exception. Datasets and NLP tools and services are made available through various repositories (institutional, disciplinary, general purpose, etc.), which makes it hard to find the appropriate resources for one's purposes. Even if they are brought together in one catalogue, such as the European Open Science Cloud\footnote{\url{https://www.eosc-portal.eu}} or the Google dataset search service\footnote{\url{https://toolbox.google.com/datasetsearch}}, the difficulty of spotting the right resources and services among thousands still remains. Metadata plays an instrumental role in solving this puzzle, as it becomes the intermediary between consumers (humans and machines) and digital resources. 

In addition, in the European Union, with the 24 official and many additional languages, multilingualism, cross-lingual and cross-cultural communication in Europe as well as an inclusive Digital Single Market\footnote{\url{https://ec.europa.eu/digital-single-market/en}} can only be enabled and firmly established through Language Technologies (LT). The boosting of the LT domain is thus of utmost importance. To this end, the European LT industry needs to be strengthened, promote its products and services, integrate them into applications, and collaborate with academia into advancing research and innovation, and bringing research outcomes to a mature level of entering the market. The European Language Grid (ELG) project\footnote{\url{https://www.european-language-grid.eu}} aims to drive forward the European LT sector by creating a platform and establishing it as the primary hub and marketplace for the LT community. The ELG is developed to be a scalable cloud platform, providing in an easy-to-integrate way, access to hundreds of commercial and non-commercial LTs for all European languages, including running tools and services as well as data resources. Discovery of and access to these resources can only be achieved through an appropriate metadata schema. We present here the ELG-SHARE schema, which is used for the description of LT-related resources shared through the ELG platform and its contribution to the project goals. 


\section{Objectives}
\label{sec:objectives}

The ELG project \cite{rehm_elgOverview_2020} aims to foster European LT by addressing the fragmentation that hinders its development; see indicatively \cite{rehm2018b,rehm2016d}. To this end, it builds a platform dedicated to the \emph{distribution and deployment of Language Resources and Technologies (LRT)}, aspiring to establish it as the primary platform and marketplace for industry-relevant LT in Europe. The \emph{promotion of LT stakeholders and activities} and growth of their visibility and outreach is also one of its goals. Together with complementary material in the portal (e.g., training material, information on events, job offerings, etc.), ELG offers a comprehensive picture of the European LT sector.

\begin{figure*}[!ht]
    \begin{center}
        \includegraphics[scale=0.7]{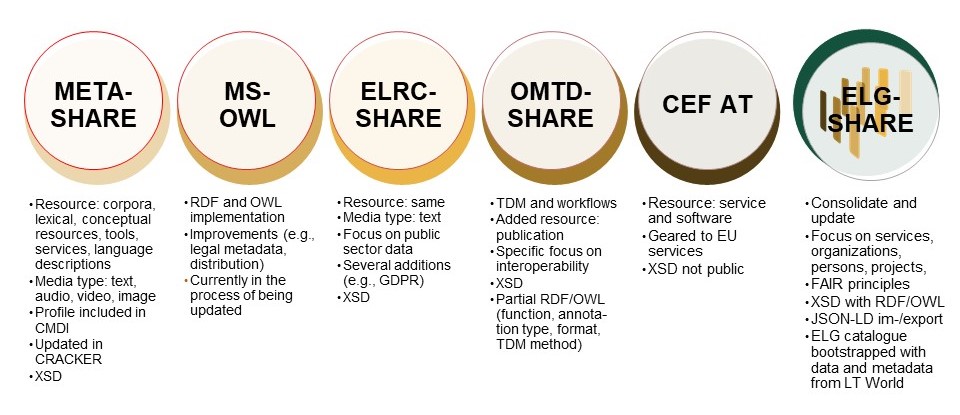} 
        \caption{Sources of the ELG metadata schema}
        \label{fig:sources}
    \end{center}
\end{figure*}

The ELG platform\footnote{The ELG platform has just been launched (alpha release) and will continue to be updated with new resources and functionalities (official release dates are on April of 2020, 2021 and 2022.} will offer access to hundreds of \emph{commercial} and \emph{non-commercial} \emph{LTs} and ancillary \emph{data LRs} for all European languages and more; these include processing and generation services, tools, applications for written and spoken language, corpora, lexicons, ontologies, term lists, models, etc. All resources are accessed through their descriptions in the ELG catalogue. LRT providers can describe, upload, and integrate their assets in ELG, and LRT consumers can download them, depending on their licensing terms, and, in the case of integrated LT services, run them through the ELG cloud platform (\emph{functional services}). The ELG catalogue includes descriptions of \emph{organizations} (companies, SMEs, academic and research organizations and groups, etc.) active in the LT sector, as well as national and European \emph{projects} related to LT. Users interested in LT can filter and search for services, data resources, organizations and more by languages, service types, domains, etc., and view their detailed descriptions. In addition, customized views (e.g., HTML pages) on the catalogue will allow users to get a summary of LT assets by specific languages, domains or application areas, and thus make knowledgeable plans for the development and exploitation of LT. 

Given its mission, ELG targets various types of users, broadly classified into (a) \emph{providers of LRTs}, both commercial and academic ones, albeit with different requirements (the former, seek to promote their products and activities, while the latter, wish to make their resources available for research or look for cooperation to further develop them in new projects or, even, commercialize them), (b) \emph{consumers of LT}, including companies developing LT tools, services and applications, integrators of LT in applications, researchers using language processing services for their studies, etc., and even (c) \emph{non-LT experts} interested in finding out more about LT and its uses. 

Last but not least, ELG is conceived as part of the emerging European ecosystem of infrastructures and initiatives that work on human-centric Artificial Intelligence (AI) research and applications. In this context, ELG aspires to be the dedicated platform that covers the special needs of the NLP/LT part of the AI community. 

The ELG platform is based on an architecture that facilitates integration, discovery and deployment of resources \cite{piperidis2019}. One of its main pillars is the metadata schema used for the formal description of all entities targeted by the ELG platform appropriately designed to meet ELG objectives. Thus, the metadata schema must:
\begin{itemize}
    \item support \emph{findability} of LT entities and facilitate \emph{accessibility} and \emph{usability} of LT assets by human users and, where possible, by machines, thus ensuring their \emph{reusability}; 
    \item enable \emph{documentation} for all types of entities \emph{at different levels of granularity}, in response to the varying user needs; for LRTs, these range from a minimum subset of information indispensable for discovery and (in the case of LTs) operation, to a rich set of properties which covers the whole lifecycle of their production and consumption and their relations to other resources and stakeholders;  for organizations, especially companies, the most detailed set includes features intended for marketing of their products and services;
    \item provide for appropriate \emph{linking among LT entities} (i.e,. across resources, as well as between resources and other entities);
    \item cater for \emph{interoperability} with other metadata schemas enabling import and export of metadata descriptions from and to collaborating platforms (cf. Section \ref{sec:methodology}).
\end{itemize}

\section{Methodology and Related Work}
\label{sec:methodology}

The ELG metadata schema (or ELG-SHARE in short) builds upon, extends and updates previous metadata works (Figure \ref{fig:sources}). Its main source is META-SHARE, a well-established and widely used schema catering for the description of LRTs in the LT domain, together with its application profiles\footnote{It should be noted that {META-SHARE} is also registered in the CLARIN Component Registry (\url{https://catalog.clarin.eu/ds/ComponentRegistry}) and used in the Greek CLARIN (\url{https://www.clarin.gr/}) and various META-SHARE nodes harvested by the Virtual Language Observatory (\url{https://vlo.clarin.eu/}).}, which adapt the core properties and relations to the needs of specific platforms  \cite{gavrilidou_meta-share_2012,mccrae_one_2015,piperidis2018,labropoulou_openminted2018}. META-SHARE was based on an extensive study of related metadata schemas and catalogues, focusing mainly on LRTs but also taking into account general trends in the metadata domain \cite{labropoulou_documentation_2012}. In the course of time, its principles and implementation policies have been updated to reflect advancements in the metadata area. 

In ELG, modifications, updates and extensions in the contents (metadata elements and values) are made in response to user requirements \cite{melnika2019} and new descriptive needs, such as:
\begin{itemize}
    \item integration and deployment of \emph{functional services} in the platform according to the ELG technical specifications \cite{rehm_elgOverview_2020},
    \item representation of licensing and billing terms for services (e.g., charging based on CPU and storage usage, etc.),
    \item more detailed description of ML models,
    \item enriched description of organizations, individuals and projects.
\end{itemize}

For the design and implementation of the ELG schema we have also taken into account user feedback from previous schemas as well as current developments in the metadata area at large, such as the FAIR principles\footnote{The FAIR principles target Findability, Accessibility, Interoperability and Reuse of digital assets, with the goal to improve data management, sharing and usage; see \url{https://www.force11.org/group/fairgroup/fairprinciples} and \url{https://www.go-fair.org/fair-principles/}.}, the Data and the Software Citation Principles\footnote{The Data Citation Principles are a set of guiding principles for citing data within scholarly literature, or any other dataset, or research object, while the Software Citation principles is a follow-up for the citation of software; see  \url{https://www.force11.org/datacitationprinciples} and \url{https://www.force11.org/software-citation-principles}.} and the DataCite schema\footnote{\url{https://schema.datacite.org/meta/kernel-4.1/doc/DataCite-MetadataKernel_v4.1.pdf}}, considerations on reproducibility of research experiments, the Open Access movement, OpenAIRE\footnote{OpenAIRE (\url{https://www.openaire.eu}) is an infrastructure dedicated to promoting and facilitating openness in scholarly literature and research; see \url{https://guidelines.openaire.eu}} guidelines for research data, and relevant RDA recommendations\footnote{The Research Data Alliance (RDA) is a community-driven initiative aiming to build the social and technical infrastructure to enable open sharing and re-use of data. The RDA endorsed outcomes can be found at  \url{https://www.rd-alliance.org/recommendations-and-outputs/all-recommendations-and-outputs}.}. All these have led to improvements in the schema contents as well as to its representation, which is currently based on OWL\footnote{\url{https://www.w3.org/OWL/}} ontologies and compatible with the Linked Data paradigm\footnote{\url{http://linkeddata.org/}}, as described in Section~\secref{sec:implementation}.

In addition, we have examined popular schemas in the area of datasets, mainly DCAT\footnote{\url{https://www.w3.org/TR/vocab-dcat} and \url{https://www.w3.org/TR/vocab-dcat-2}} and schema.org\footnote{\url{https://schema.org/Dataset}}, ensuring that ELG metadata records can be exported in a compliant form to other popular distribution catalogues and, thus, ensuring a wider uptake of the LT products and services included in ELG. 

Finally, we have initiated collaborations with neighbouring initiatives and projects leading to a "common pool of resources" that communities can share, adapt and exploit to their respective needs. Interoperability is a key issue in this endeavour and the first level relates to metadata. Crosswalks between the minimal schema of ELG and the schemas of other projects has started with the ontology used for the description of AI resources in AI4EU\footnote{\url{https://www.ai4eu.eu}}\cite{rehm2020}.
Given the flexibility of CMDI\footnote{\url{https://www.clarin.eu/content/component-metadata}} \cite{BROEDER12.581}, which is the metadata framework adopted in CLARIN\footnote{\url{https://www.clarin.eu/}}, and the fact that the ELG schema builds upon META-SHARE, which is already included among the CMDI profiles, the exchange of metadata records between the two catalogues can easily be established.

Finally, for the enriched descriptive modules of LT stakeholders and activities, we have explored various schemas and ontologies, such as FOAF\footnote{\url{http://xmlns.com/foaf/spec/}} for persons and organizations, the LT-Innovate catalogue of LT actors\footnote{\url{http://www.lt-innovate.org/directory}}, BIBO\footnote{\url{http://bibliontology.com}} and OpenAIRE for bibliographic records, and more. The LT-World \cite{jorg2010}, a (no longer online) ontology-driven web portal aimed at serving the global LT community and providing information on organizations, projects, events, resources, products, etc. in the LT domain, has also been considered, while part of its data will be used to bootstrap the catalogue. 

This approach, of building on widespread metadata schemas by adapting, updating and enriching them, empowers the re-use of an initial set of metadata records from platforms and catalogues (e.g., META-SHARE\footnote{\url{http://www.meta-share.org}}, ELRC-SHARE\footnote{\url{https://elrc-share.eu}}, ELRA catalogue\footnote{\url{http://catalogue.elra.info/en-us/}}, etc.) through an easy conversion process. It also facilitates the adoption of the new schema by LRT providers who are already familiar with the source schemas. Furthermore, the adoption of the Linked Data paradigm ensures interoperability with external catalogues and enhances the role of ELG as an LT supplier for other communities.

\section{Presentation of the Schema}
\label{sec:schema}
\subsection{ELG Entities}
\label{sec:entities}
ELG-SHARE is the backbone of the ELG platform, as it supports the registration and discovery of all entities and facilitates the operation of functional services. 
It aims to formalize the description of \textbf{language processing tools/services} and the \textbf{data resources} that are required for their operation and development, such as models, ontologies and term lists that can be used as ancillary resources at processing time, or corpora that can be used for training. More specifically, the ELG schema brings together under the term \textbf{language resource}\footnote{The term "Language Resource" is used mainly for resources composed of linguistic material used in the construction, improvement or evaluation of language processing applications, but also, in a broader sense, in language and language-mediated research studies and applications. The term is often used in the bibliography and related initiatives with a broader meaning, encompassing also (a) tools and services used for the processing and management of datasets, and (b) standards, guidelines and similar documents that support the research, development and evaluation of LT. In the ELG schema, we use the term as first defined in  META-SHARE, i.e., including both data resources and LTs.} the following:
\begin{itemize}
\item \emph{tools and services}, including any type of software that performs language processing and/or any LT-related operations (e.g., annotation, machine translation, speech recognition, speech-to-text synthesis, visualization of annotated datasets, training of corpora, etc.); 
\item \emph{corpora and datasets}, defined for our purposes as structured collections of pieces of language data typically of considerable size and selected according to criteria external to the data (e.g., size, language, domain, etc.) to represent as comprehensively as possible a specific object of study;
\item \emph{lexical and conceptual resources}, i.e., resources such as term glossaries, word lists, semantic lexica, ontologies, etc., organized on the basis of lexical or conceptual units (lexical items, terms, concepts, phrases, etc.) with their supplementary information (e.g., grammatical, semantic, statistical information, etc.);
\item \emph{language descriptions}, which include resources aiming to model a language or some aspect(s) of a language via a systematic documentation of linguistic structures; examples in this category include statistical and machine learning-computed language models and computational grammars.
\end{itemize}

In addition, the schema caters for the description of \textbf{related/satellite entities} that are involved in the lifecycle of LRTs:
\begin{itemize}
\item \emph{actors}, i.e., \emph{organizations, groups or persons}, who have created or distribute a resource, act as contact persons, participate in a project, etc.;
\item \emph{projects} that have funded the creation, maintenance or extension of a resource, or in which a resource may have been used; 
\item \emph{documents}, such as installation and user manuals of  a tool, publications of a research experiment where a resource has been used, etc.; and
\item \emph{licences/terms of use} regulating the use of LRTs.
\end{itemize}

In META-SHARE and its application profiles, only a small set of features were suggested for the description of these entity types. In ELG, they play a more central role and, thus, their metadata modules have been extended to accommodate the project objectives, as described in \ref{sec:structure}

\begin{figure}[!ht]
    \begin{center}
        \includegraphics[scale=0.25]{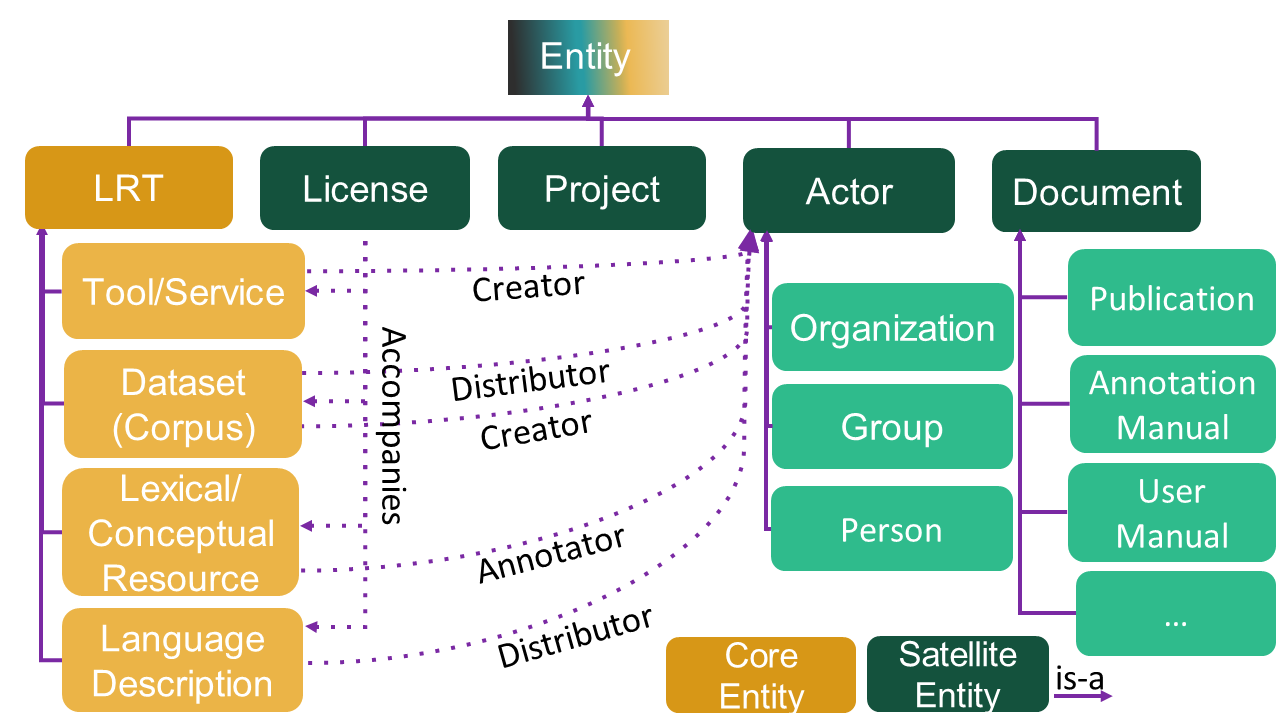} 
        \caption{ELG entities}
        \label{fig:entities}
    \end{center}
\end{figure}

Figure~\ref{fig:entities} shows a conceptual, hierarchical representation of the entities described by the ELG metadata schema and exemplary relations among them.

\subsection{Describing LRTs}
\label{sec:structure}
The schema caters for their full lifecycle, from conception and creation to integration in applications and usage in projects, also recording relations with other resources 
(e.g., raw and annotated versions of corpora, tools used for their processing, models integrated in tools, etc.) and related/satellite entities (see Section~\ref{sec:entities}).

To encode this wealth of information, the ELG schema includes a large number of metadata elements grouped along three key concepts: \textbf{resource type}, \textbf{media type} and \textbf{distribution}.
The \textbf{resource type} element distinguishes LRs in the four classes presented in Section~\ref{sec:entities}
\textbf{Media type} refers to the form/physical medium of a data resource (or of its parts, in the case of multimodal resources), i.e., ~\emph{text, audio, image, video} and \emph{numerical text} (used for biometrical, geospatial and other numerical data).
Finally, \textbf{distribution}, following the DCAT vocabulary, refers to the physical form of the resource that can be distributed and deployed by consumers; for instance, software resources may be distributed as web services, executable files or source code files, while data resources as PDF, CSV or plain text files or through a user interface. Administrative and descriptive metadata are mostly common to all LRTs, while technical metadata differ across resource and media types as well as distributions. 

In the first instance, this abundance of information in the schema makes tedious the process of creating metadata records. To ensure flexibility and uptake, metadata elements are distinguished into \emph{mandatory}, \emph{recommended} and \emph{optional} ones. This allows us to set up a \textbf{minimal version} through a careful selection of mandatory and strongly recommended elements. The same approach has been used in the predecessors of the ELG schema, but each time the selection was adjusted to the platform objectives. For ELG, the criteria used include: required for \emph{discovery}, especially features considered of high interest to ELG consumers \cite{melnika2019b}; considered indispensable for \emph{accessing} the resources and, in the case of functional services, ensuring proper \emph{deployment} in the ELG infrastructure; supporting \emph{usage} of the resources; deemed valuable for \emph{research experiments and projects} and essential for achieving \emph{interoperability} with metadata used for the platforms of the broader communities. 

In this way, the population of the ELG platform can follow a staged approach, whereby metadata records are initially created with only the minimal information and then gradually enriched, e.g., through various manual and (semi-)automatic processes. 
It also makes easier the population of the platform by harvesting processes from other sources (catalogues, repositories, etc.) which may host metadata records with less information. In this scenario, the metadata creators themselves, and (in the case of harvesting) assigned persons from the consortium or individuals who "claim" a metadata record will have the chance to  curate, further enrich and validate its metadata. 

The minimal version of the ELG schema includes the following metadata categories of information: 
\begin{itemize}
    \item for \emph{all types of resources}: resource names; identifiers; a short description of its contents; versioning data; a point for further information (email or landing page); information on the metadata record itself (e.g., data of the metadata editor or harvesting source, creation date, etc.); data of the resource provider; classification by domain and keywords; links to manuals, training material, samples of the resource; licensing conditions, access location and form for each distribution of the resource;
    \item for \emph{tools/services}: service/application type; specifications for the input resource that a tool can process with regard to languages, media type and formats; information on the output resource, again for languages, media type and formats, as well as annotation/extraction types (e.g., lemmas, named entities, sentiment tags, etc.); hardware/software requirements (e.g., RAM); links to the ancillary resources (e.g., models, lexica, word lists, etc.) used at operation; for \emph{functional services}, docker image location and execution endpoint;
    \item for\emph{ all data LRs}: language coverage; size and formats per distribution; 
    \item for \emph{corpora and datasets}: classification elements, which may be media-dependent (e.g., audio genre, text type, etc.); if they are processed, information at least for the annotation types, and link to the raw version;
    \item for \emph{lexical/conceptual resources}: subtype (e.g. ontology, lexicon, etc.); meta-language; basic unit of description (i.e., lemma, concept, etc.); types of the accompanying linguistic or extra-linguistic information (e.g., part-of-speech tags, senses, translation equivalents, etc.); 
    \item for \emph{language descriptions}: subtype (e.g., grammar, model); meta-language; types of linguistic or extra-linguistic information; for models, information on the training corpus and the framework.
\end{itemize}

Optional metadata categories record, for instance, resource/media-independent creation details (e.g., resource creators, funding projects), and  media-dependent ones (e.g., related to the recording process of videos and audios), information on the projects and applications where the resource has been used. 

We should note here that the schema includes features that enable \emph{interoperability across resource types}. These are important for enhancing the functionalities of the ELG platform as well as for future extensions and collaborations with other platforms\cite{rehm2020}. Thus, \emph{format} and \emph{language} can be used to match together tools/services with candidate input resources and initiate their processing; for instance, a tool that takes as input PDF files can be matched with datasets in PDF format. Similar information can also be used to semi-automatically compose workflows of tools and/or match together tools with compatible ancillary resources (annotation resources, ontologies, ML models) to create services and end-user applications \cite{piperidis2015,labropoulou_openminted2018}. 

\begin{figure}[!ht]
    \begin{center}
        \includegraphics[scale=0.20]{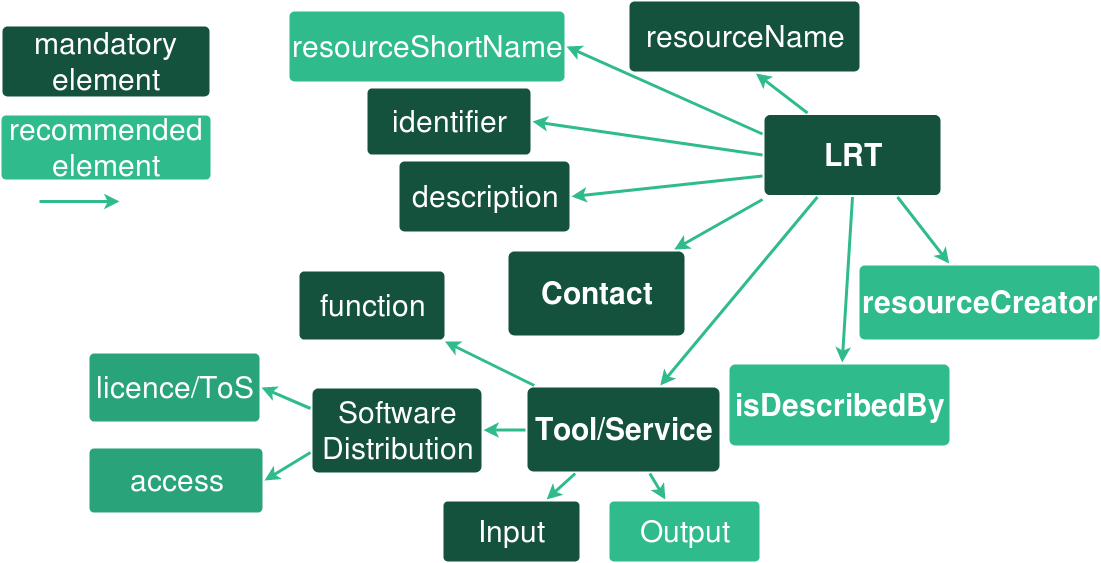} 
        \caption{Simplified subset of the ELG metadata schema}
        \label{fig:schema_visualization}
    \end{center}
\end{figure}

Figure~\ref{fig:schema_visualization} shows a simplified subset of the metadata schema with its structuring layers and optionality status. 

\subsection{Describing LT-related entities}
ELG intends to offer the \emph{who is who} of \emph{actors} and \emph{projects} in LT. Thus, the module for actors and projects, in comparison to META-SHARE profiles, has been enriched. Besides identification and descriptive metadata elements, such as name/title, identifier(s), contact information, etc., of particular importance are features related to and/or promoting LT activities, products and services, such as links to LRTs, logos, promotional material, specialization area of LT or domain, etc. 

\emph{Documents} related to LRTs (e.g., user manuals, publications, etc.) are described with mainly bibliographic metadata and, optionally, a category of the LT area to which they belong.

\emph{Licences and terms of use} are described by a set of mainly administrative metadata (e.g., licence name, access URL) and elements facilitating human users to understand the main access conditions \cite{rodriguez-doncel_digital_2015}. The module will also include a set of information for billing requirements of commercial services (currently work in progress).

\section{Language Technology Taxonomy}
\label{sec:taxonomy}
For standardization purposes, the ELG schema, in line with META-SHARE principles \cite{piperidis2012}, favours controlled vocabularies over free-text fields, especially when these are associated with internationally acknowledged standards, best practices or widespread vocabularies (e.g., ISO 3166 for region codes, RFC 5646 for languages, etc.). Specially devised vocabularies are used for various metadata elements, mainly for features specific to the LT sector. One such prominent case is the \textbf{LT application area}.

\begin{figure}[!ht]
    \begin{center}
        \includegraphics[scale=0.75]{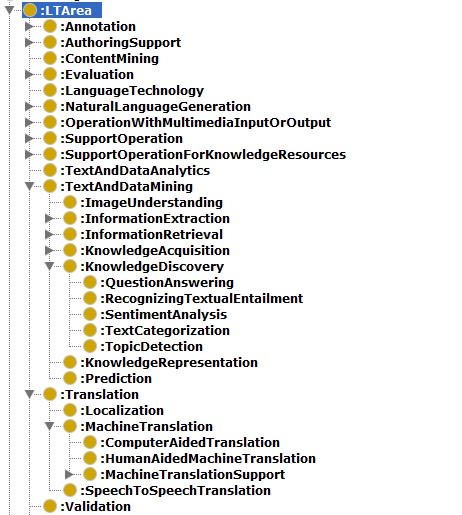} 
        \caption{Excerpt of the Language Technology taxonomy}
        \label{fig:lt_taxonomy}
     \end{center}
\end{figure}

The 'LT application area' element is the main linking bridge between all entities in the ELG catalogue. It is used, for instance, to classify LTs by the function/task they perform ('service/application type'), data LRs with respect to the applications they are intended for or have been used for, organizations by the area they are active in, etc. Its values are drawn from a hierarchically structured vocabulary, referred to as \emph{"LT taxonomy"} (Figure \ref{fig:lt_taxonomy}). The platform will also offer customised views of its contents based on the LT taxonomy in the form of a catalogue, for instance, of all actors involved in a certain LT area, of the LT area with the largest number of tools/services or companies, etc. This functionality helps raise awareness and promote LT among the field experts, by providing an overview of the LT activities in relation to various criteria.

Like all controlled vocabularies in ELG, the LT taxonomy is formally represented as an OWL class (cf. Section~\secref{sec:implementation}). This has a number of benefits as discussed in Section \ref{sec:usage}


Work on the LT taxonomy started for the OMTD-SHARE profile \cite{labropoulou_openminted2018} and included a systematic curation of free-text values used in META-SHARE metadata records, but focusing on Text and Data Mining areas. In ELG, we have extended it to cover broader LT domains and operations, incorporating feedback from all consortium partners and collaborating projects. We aim to continue extending it to cover further needs and new emerging areas. Various ways of enriching it are envisaged: as a minimum, following a process that evaluates relevant values added in the 'keyword' element by LRT providers as candidates for inclusion in the taxonomy.

\section{Implementation Issues}
\label{sec:implementation}

The ELG metadata schema is formally described with the XML Schema Definition (XSD) language. Its elements are linked to entities from two ontologies, namely the META-SHARE and the OMTD-SHARE ontology\footnote{The new versions for both ontologies have been pre-released at \url{http://purl.org/net/def/metashare} and \url{http://w3id.org/meta-share/omtd-share/} respectively.}. Specifically, each metadata element and value has an identifier which contains the IRI of the corresponding entity. 

The proposed approach contributes to the FAIRness of the metadata in ELG, makes easier their linking to metadata records in other catalogues, and supports import/export in the JSON-LD serialization format, which increases the visibility of ELG assets overall. On the other hand, the use of the XSD enables us to transform the metadata schema into the form of an entity-relationship model, facilitating its documentation and its conversion into the ELG catalogue backend relational database. XML metadata records compliant with the schema can also be imported/exported (see figure~\ref{fig:xmlrecord} in the Appendix).

The conversion of the ontology entities into XML elements has been automatically performed using a python script, thus enabling an easy update of the schema alongside the ontology update. In addition, all relations, labels and definitions are copied into the XML elements with the same script. 

ELG supports the management of metadata records through various mechanisms, facilitating LRT providers, novice and expert, when integrating their assets in the platform, and consumers when they search for, view and deploy the available LRTs. Metadata editing forms integrated in the ELG catalogue GUI will allow users to create, edit and delete metadata records, while a batch metadata import service will also be available. All different mechanisms for populating the ELG catalogue, including the harvesting service on the basis of agreed  protocols, communicate with the catalogue backend through a REST API, which returns JSON files compliant with the schema \cite{piperidis2019}. Validation services will be implemented providing meaningful messages for non-compliant input JSON files. The same API contains endpoints for exporting metadata records. Metadata converters from and to popular schemas will also be made available through the ELG portal. Finally, user guidelines and tutorials (face-to-face and short videos) will be created for documenting the use of the schema. 

Functionalities for managing the metadata schema itself are foreseen for administrators only. For instance, statistics on the use of metadata elements and values will be used to evaluate their uptake and taken into account for the schema updates.

\section{Deployment of the Metadata Schema in ELG}
\label{sec:usage}

\subsection{Metadata as Catalogue Filters}
The ELG catalogue is an important asset that benefits the different stakeholders in the NLP, Knowledge Management and broader LT industry, including users of the technology itself, LT vendors and technology integrators, as well as academic institutions and research communities active in the field. The centralized repository, a comprehensive base of functional services, language-specific models and other resources (including those for less-resourced languages), the rich set of metadata and controlled vocabularies, in connection with the mechanisms put in place to guarantee data interoperability and to enhance browsing, searching and discovering information transforms the catalogue into a unique resource that can boost LT research and industry in Europe and beyond.

The catalogue enables users to find tools, services and data resources thanks to the metadata elements and controlled vocabularies offering different facets to narrow the search to specific service types (e.g. sentiment analysis), supported languages or licences. In addition, users can take a glance at the LT landscape of tools and resources by browsing the facets in the LT taxonomy or following an exploratory search approach. This allows them to discover new services or resources they had not been aware of or to identify alternative services to the ones they already use or plan to use. The broad information contained in the catalogue facilitates users when choosing tools and resources for each project.  

Users also benefit from the competition emerging between LT providers listed in the catalogue aiming to keep or increase their market share and trying to outpace their competitors in terms of reliability, support and price. 

LT providers, on the other hand, gain market visibility when adding their services to the catalogue. In addition, the catalogue enables LT providers and integrators to locate complementary services allowing them to tackle more complex projects which otherwise could not be implemented easily. Companies can use the information in the catalogue for business intelligence purposes and perform market and competitors' analyses. This information can be employed to devise strategies in order to increase market share and adjust the portfolio of services and prices in the catalogue. Furthermore, system integrators and consultancy firms can find and link-up with potential partners who provide the exact expertise and experience required for a particular project. These can be identified based on the catalogue information describing providers, their services, supported technologies and languages, licensing and pricing strategies.

In the academic field, the ELG platform is expected to positively influence how new tools and datasets emerging from research projects are shared, reused and reproduced. Researchers can not only register in the catalogue the tools that have been produced as a part of their research projects but also enter the data used to validate and support their findings. Thus, the catalogue will contain sufficient information to allow reproducibility of scientific findings. In addition, the catalogue will serve as a fundamental tool to monitor and survey the state of the art about LT services including scientific contributions and commercial products.

\subsection{Metadata and the Data Management Plan}
\label{sec:dmp}
A key element in the lifecycle of a LR, its proper management and its long-term sustainability implies following the guidance of a Data Management Plan (DMP). This has been reinforced by Article 29.3 of the H2020 Grant Agreement, which has made the implementation of a DMP a prerequisite for any H2020 submission that makes use of data. Such DMP must comply with the FAIR principles defined in the H2020 Participant Portal manual.
As part of ELG’s tasks, a DMP procedure is being implemented to ensure that for all LRTs that are collected/produced, packaged and shared/repurposed within ELG and its Pilot Projects, all required information is included to help identify all issues having an impact on the data collection and description (metadata) processes. A first version of the DMP has been produced in June 2019 \cite{kamocki2019} and  updated in December. This document provides the guidelines to be followed within ELG on the overall data management lifecycle. It also includes a template drafted specifically for LRs to provide accurate information on the different steps carried out during their lifecycle.  These steps are distributed over three main tasks: (1) Data Acquisition: covering both pre-existing and new LRs (with their production phase and validation steps), as well as the post-production phase (with licences, allocation of unique identifiers (PID, DOI, ISLRN), documentation, etc.); (2) Storage, Preservation and Access: considering all aspects that will have an impact on the future sharing of the LRTs, such as physical storage and backups, allowing for their potential customization and/or improvement, ensuring data integrity and confidentiality (e.g., whether data have been anonymized); (3) Sharing: describing availability, access restrictions (if any), licences and rights to share.
In order to make LRs included in the ELG platform FAIR, the DMP template will be linked to the metadata schema so that each LR produced, thanks to ELG support or through conversion of existing resources, is appropriately described in the catalogue.

\section{Conclusions and Future Work}
\label{sec:conclusions}
The first release of the ELG platform will be made available in March 2020 and . The first release will include v1 of the schema and updated versions will be made available with the next releases. 

The first release of the ELG platform (launched in March 2020) - to be followed by two more releases, in February and September 2021 - is built with schema v1.1\footnote{The schema XSD (continuously updated), documentation and exemplary metadata records for it are available at: \url{https://gitlab.com/european-language-grid/platform/ELG-SHARE-schema/} licensed under CC-BY-4.0.}. The platform includes metadata records for more than 300 data resources available with open licenses that have been selected and converted from  three catalogues (ELRA, ELRC-SHARE and META-SHARE), while harvesting from LINDAT-CLARIAH(CZ)\footnote{https://lindat.mff.cuni.cz}  is in the immediate plans. Around 170 functional LT services have been manually described by the consortium partners and a smaller set of records for projects and organizations related to LT converted from other sources (e.g. the EU open data portal). Feedback from the creators of these metadata and, most important, of the platform users will be taken into account for future improvements of the schema and the ontologies. Valuable input will also be provided by collaborating projects (e.g. \emph{ICT-29-208 subtopic b) projects}). 
 
Ongoing work is focusing on the billing module for commercial services; we have started discussions on the specifications which will be formalised in the schema. 
Functionalities for supporting the metadata schema (metadata editor, automatic metadata enrichment, curation of the metadata schema, etc.) are also in our plans.

\section{Acknowledgements}
Work reported in this paper has been carried out in the framework of the ELG project, which has received funding from the European Union's Horizon 2020 research and innovation programme under grant agreement no.~825627. We would like to thank all our colleagues in the ELG consortium for their contributions to the metadata schema. We would also like to thank all those who have contributed to previous versions of the schema and ontologies. 

\section{Bibliographical References}
\label{main:ref}

\bibliographystyle{./lrec}
\bibliography{./metadataSchema}


\clearpage

\definecolor{maroon}{rgb}{0.5,0,0}
\definecolor{darkgreen}{rgb}{0,0.5,0}
\lstdefinelanguage{XML}
{
  basicstyle=\ttfamily\scriptsize,
  morestring=[s]{"}{"},
  morecomment=[s]{?}{?},
  morecomment=[s]{!--}{--},
  commentstyle=\color{darkgreen},
  moredelim=[s][\bfseries\color{black}]{>}{<},
  moredelim=[s][\bfseries\color{red}]{\ }{=},
  stringstyle=\color{blue},
  identifierstyle=\bfseries\color{maroon}
}


\begin{figure*}[h]
\centerline{\large\textbf{Appendix}}
\lstset{language=XML}
\begin{lstlisting}
<ms:MetadataRecord>
  <ms:MetadataRecordIdentifier ms:MetadataRecordIdentifierScheme="ms:elg">ELG_MDR_LTS_291119_00000002
  </ms:MetadataRecordIdentifier>
  <ms:metadataCreationDate>2019-11-29</ms:metadataCreationDate>
  <ms:metadataLastDateUpdated>2019-11-29</ms:metadataLastDateUpdated>
  <ms:metadataCurator>
    <ms:surname xml:lang="en">Smith</ms:surname>
    <ms:givenName xml:lang="en">John</ms:givenName>
  </ms:metadataCurator>
  <ms:compliesWith>ms:ELG-SHARE</ms:compliesWith>
  <ms:metadataCreator>
    <ms:surname xml:lang="en">Smith</ms:surname>
    <ms:givenName xml:lang="en">John</ms:givenName>
  </ms:metadataCreator>
  <ms:DescribedEntity>
    <ms:LanguageResource>
      <ms:entityType>languageResource</ms:entityType>
      <ms:resourceName xml:lang="en">ANNIE English Named Entity Recognizer</ms:resourceName>
      <ms:resourceShortName xml:lang="en">ANNIE</ms:resourceShortName>
      <ms:description xml:lang="en">Named entity recognition pipeline that identifies ...</ms:description>
      <ms:LRIdentifier ms:LRIdentifierScheme="ms:elg">ELG_ENT_LTS_291119_00000035</ms:LRIdentifier>
      <ms:version>8.6</ms:version>
      <ms:additionalInfo>
        <ms:landingPage>https://cloud.gate.ac.uk/...</ms:landingPage>
      </ms:additionalInfo>
      <ms:contact>
        <ms:Person>
          <ms:surname xml:lang="en">Smith</ms:surname>
          <ms:givenName xml:lang="en">John</ms:givenName>
        </ms:Person>
      </ms:contact>
      <ms:keyword xml:lang="en">GATE</ms:keyword>
      <ms:keyword xml:lang="en">NER</ms:keyword>
      <ms:keyword xml:lang="en">English</ms:keyword>
      <ms:resourceProvider>
        <ms:Organization>
          <ms:organizationName xml:lang="en">University of Sheffield</ms:organizationName>
        </ms:Organization>
      </ms:resourceProvider>
      <ms:validated>false</ms:validated>
      <ms:LRSubclass>
        <ms:ToolService>
          <ms:lrType>toolService</ms:lrType>
          <ms:function>ms:NamedEntityRecognition</ms:function>
          <ms:function>ms:PosTagging</ms:function>
          <ms:SoftwareDistribution>
            <ms:SoftwareDistributionForm>ms:dockerImage</ms:SoftwareDistributionForm>
          </ms:SoftwareDistribution>
          <ms:digest>c107...</ms:digest>
          <ms:downloadLocation>https://registry.gitlab.com/...</ms:downloadLocation>
          <ms:additionalHwRequirements>none</ms:additionalHwRequirements>
          <ms:LicenceTerms>
            <ms:licenceTermsName>LGPL-3.0-only</ms:licenceTermsName>
          </ms:LicenceTerms>
          <ms:languageDependent>TRUE</ms:languageDependent>	
          <ms:inputContentResource>
            <ms:processingResourceType>ms:file1</ms:processingResourceType>
            <ms:languageTag>en</ms:languageTag>
            <ms:mediaType>text</ms:mediaType>
            <ms:dataFormat>ms:Text</ms:dataFormat>
            <ms:dataFormat>ms:Html</ms:dataFormat>
          </ms:inputContentResource>
          <ms:outputResource>
            <ms:processingResourceType>ms:file1</ms:processingResourceType>
            <ms:languageTag>en</ms:languageTag>
            <ms:mediaType>text</ms:mediaType>
            <ms:annotationType>ms:Date</ms:annotationType>
            <ms:annotationType>ms:Organization</ms:annotationType>
            <ms:annotationType>ms:Person</ms:annotationType>
            <ms:annotationType>ms:Location</ms:annotationType>
          </ms:outputResource>
        </ms:ToolService>
      </ms:LRSubclass>
    </ms:LanguageResource>
  </ms:DescribedEntity>
</ms:MetadataRecord>
\end{lstlisting}
    \caption{Example of a metadata record for a functional service}
    \label{fig:xmlrecord}
\end{figure*}

\end{document}